\documentclass[conference]{IEEEtran}

\usepackage{courier}
\usepackage{amstext}
\usepackage{times}
\usepackage{helvet}
\usepackage{courier}
\usepackage[cmex10]{amsmath}
\usepackage{amssymb}
\usepackage{latexsym}
\usepackage{eurosym}
\usepackage{wasysym}

\makeatletter
\newif\if@restonecol
\makeatother

\usepackage{etex}
\usepackage{tikz}
\usepackage{pslatex}
\usepackage[arrow, matrix, curve]{xy}
\usepackage{epsfig}
\usepackage{pstricks,egameps}

\begin{document}

\title{Reciprocity in Gift-Exchange-Games}

\author{\IEEEauthorblockN{Rustam Tagiew}
\IEEEauthorblockA{Alumni of\\TU Bergakademie Freiberg\\
yepkio@mail.ru}
\and
\IEEEauthorblockN{Dmitry I. Ignatov}
\IEEEauthorblockA{National Research University\\
Higher School of Economics\\
dignatov@hse.ru}}

\maketitle

\begin{abstract}
This paper presents an analysis of data from a gift-exchange-game experiment. The experiment was described in `The Impact of Social Comparisons on Reciprocity' by G\"achter et al. 2012. Since this paper uses state-of-art data science techniques, the results provide a different point of view on the problem. As already shown in relevant literature from experimental economics, human decisions deviate from rational payoff maximization. The average gift rate was $31$\%. Gift rate was under no conditions zero. Further, we derive some special findings and calculate their significance.
\end{abstract}

\IEEEpeerreviewmaketitle

\section{Introduction}
\indent It is not only the global financial crisis of the recent years \cite{maxton}, which made economists reconsider the path economics as a discipline should take. Since decades, it became obvious that pure theories fail in real world \cite{ariely}. Paul Krugman described the current situation in economics as: `... the central cause of the profession’s failure was the desire for an all-encompassing, intellectually elegant approach that also gave economists a chance to show off their mathematical prowess. Unfortunately, this romanticized and sanitized vision of the economics led most economists to ignore all the things that can go wrong. They turned a blind eye to the limitations of human rationality'. The promising solution to that problem -- experimental economics -- gained at importance. Human subject research shifted economists' point of view closer to the psychologists' one -- people are no more considered to be rational payoff maximizers. On the other side, growing size and complexity of the data 
makes the application of state-of-art data science essential.\\
\indent Before vast data and computational power were available, classical economists used game theory to predict outcomes of human interactions. People were assumed to be intelligent and autonomous, and to act pursuant to their existing preferences. It is important to underline that game theory is a mathematical discipline, whose task was never to define human preferences, but to calculate based on their definition. A preference is an order on outcomes of an interaction. One can be regarded as rational, if one always makes decisions, whose execution has referred to subjective estimation the most preferred consequences \cite{russel,rubin}. The level of intelligence determines the correctness of subjective estimation. Beyond justifying own decisions, rationality is a base for predictions of other people's decisions. If the concept of rationality is satisfied, and applied mutually, and even recursively in a human interaction, then the interaction is called strategic. Game is a notion for the formal structure 
of a concrete strategic interaction \cite{neumann}.\\
\indent A definition of a game consists of a number of players, their preferences, their possible actions and the information available for the actions. A payoff function can replace the preferences under assumed payoff maximization. The payoff function defines each player's outcome depending on his actions, other players' actions and random events in the environment. The game-theoretic solution of a game is a prediction about the behavior of the players also known as an equilibrium. The basis for an equilibrium is the assumption of rationality. Deviating from an equilibrium is outside of rationality, because it does not maximize the payoff according to the formal definition. There are games, which have no equilibria. At least one mixed strategies equilibrium is guaranteed in finite games \cite{nash}.\\
\indent In common language, the notion of game is used for board games or video games. In game-theoretic literature, it is extended to all social, economical and pugnacious interactions among humans. A war can be simplified as a board game. Some board games were even developed to train people, like Prussian army war game `Kriegspiel Chess' \cite{wargaming} for their officers. We like it to train in order to perform better in games \cite{ggp}. In most cases, common human behavior in games deviates from game-theoretic predictions \cite{poolgame,vspt}. One can say without any doubt that if a human player is trained in a concrete game, he will perform close to equilibrium. But, a chess master is not necessarily a good poker player and vice versa. On the other side, a game-theorist can find a way to compute an equilibrium for a game, but it does not make a successful player out of him. There are many games we can play; for most of them, we are not trained. That is why it is more important to investigate our
behavior while playing general games than playing a concrete game on expert level.\\
\indent Although general human preferences are a subject of philosophical discussions \cite{humannat}, game theory assumes that they can be captured as required for modeling rationality. Regarding people as rational agents is disputed at least in psychology, where even a scientifically accessible argumentation exposes the existence of stable and consistent human preferences as a myth \cite{sociokritik}. The problems of human rationality can not be explained by bounded cognitive abilities only. `... people argue that it is worth spending billions of pounds to improve the safety of the rail system. However, the same people habitually travel by car rather than by train, even though traveling by car is approximately 30 times more dangerous than by train!'\cite[p.527--530]{hrational} Since the last six decades nevertheless, the common scientific standards for econometric experiments are that subjects' preferences over outcomes can be insured by paying differing amounts of money \cite{performmoney}. However, 
insuring preferences by money is criticized by tossing the term `Homo Economicus' as well.\\
\indent The ability of modeling other people's rationality and reasoning as well corresponds with the psychological term `Theory of Mind' \cite{verbrugge}, which lacks almost only in the cases of autism. For experimental economics, subjects as well as researchers, who both are supposed to be non-autistic people, may fail in modeling of others' minds anyway. In Wason task at least, subjects' reasoning does not match the researchers' one \cite{wason}. Human rationality is not restricted to capability for science-grade logical reasoning -- rational people may use no logic at all \cite{nonlogic}. However, people also make serious mistakes in the calculus of probabilities \cite{tversky}. Even in mixed strategy games, where random behavior is of a huge advantage, the required sequence of random decisions can not be properly generated by people \cite{randomexp}. Due to bounded cognitive abilities, every human `random' decision depends on previous ones and is predictable in this way. In ultimatum games \cite[S. 43ff]
{vspt}, the former economists' misconception of human preferences is revealed -- people's minds value fairness additionally to personal enrichment. Our minds originated from the time, when private property had not been invented and social values like fairness were essential for survival.\\
\indent From the view point of data scientists fascinated by human behavior, the sizes of datasets originated from social networks predominate the ones from experimental economics by orders of magnitude \cite{vspt}. Nevertheless, analyzing data from experimental economics has the same importance for understanding human psychology as studying Escherichia for understanding human physiology. Data from experimental economics has the advantage of originating from simple and controlled human interactions.\\
\indent In current experimental economics, the models are first constructed by philosophical plausibility considerations and then are claimed to fit the data. In this work, we reverse the order of common research in experimental economics. We follow the slogan `existence precedes essence' -- the philosophical plausibility considerations follow after the correlations and regularities are found. For these needs, we analyze the dataset of the paper ``The Impact of Social Comparisons on Reciprocity'' by G\"achter et al. \cite{sefton}. The only assumption about human behavior is its determinism.\\
\indent The next section summarizes related work on data mining approaches and economical models. Then, the experiment setup and the gathered data are introduced. Before extracting rules of behavior, we explain the reasons for the assumption of determinism. We also explain conceptual problems of using linear model on this data. The results and their interpretations follow afterwards. Then, a section is devoted to p-hacking. A suggestion for more efficient research on human behavior is made in future work. Summary and discussion conclude this paper.\\
\section{Related Work}
\indent A similar approach is already explored on three datasets -- a zero-sum game of mixed strategies, an ultimatum game and repeated social guessing game \cite{tagieweeml,tagieweeml2}. For these datasets, extracted deterministic regularities outperformed state-of-art models. It was shown that some regularities can be easily verbalized, what underlines their plausibility.\\
\indent A very comprehensive gathering of works in experimental psychology and economics on human behavior in general games can be found in \cite{experimental}. Quantal response equilibrium became popular as a model for deviations from equilibria \cite{qre}. It is a parametrized shift between mixed strategies equilibrium and an equal distribution. The basic idea for quantal response equilibrium is the concept of trembling hand -- people make mistakes with certain probability. Unfortunately, the Akaike information criterion \cite{akaike} is rarely calculated to judge the trade-off between fit quality and model complexity \cite{tagiewPhD}. Another popular model is the linear regression. It is used in the original paper to model the dataset \cite{sefton}. For linear regression, data is translated into real numbers.\\
\section{Gift-Exchange-Game}
\begin{figure}
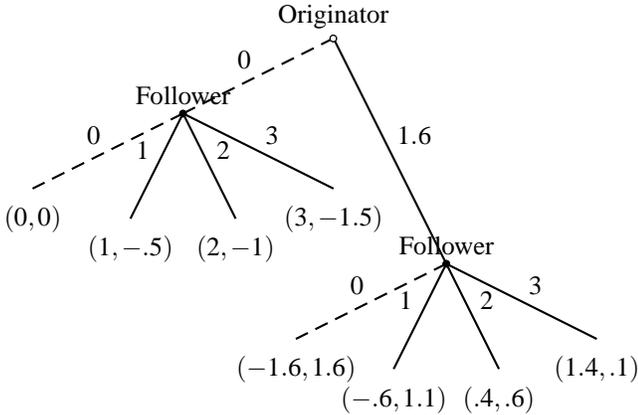

\begin{center}
\begin{egame}(600,600)
\putbranch(300,460)(-2,1){200}
\ib[linestyle=dashed]{Originator}{$0$}
\initialtrue
\putbranch(300,460)(1,2){150}
\ib{}{$1.6$}
\putbranch(100,360)(2,1){200}
\iib[linestyle=dashed][]{Follower}{$0$}{$3$}[$(0,0)$][$(3,-1.5)$]
\putbranch(100,360)(1,2){70}
\iib{}{$1$}{$2$}[$(1,-.5)$][$(2,-1)$]
\putbranch(450,160)(2,1){200}
\iib[linestyle=dashed][]{Follower}{$0$}{$3$}[$(-1.6,1.6)$][$(1.4,.1)$]
\putbranch(450,160)(1,2){70}
\iib{}{$1$}{$2$}[$(-.6,1.1)$][$(.4,.6)$]
\end{egame}
\end{center} 
\caption{Experimental non-zero-sum 2-players GEG in extensive form. (Originator's payoff, Follower's payoff) -- payoffs are in \pounds. Payoff maximizing equilibrium is marked by dashed lines.}\label{gamestr}
\end{figure}

\indent Since Akerloff and Yellen published their leading work \cite{yellen} on unemployment, gift-exchange-games (GEG) became standard for modeling labor relations. Such a game involves at least two players -- an `employer' and an `employee'. The `employer' has to decide first, whether to award a higher salary or not. Then, the `employee' has to decide, whether to put extra effort or not. Unfortunately, the experiment conducted by G\"achter et al. did not implement a real-effort task. The `employee' does not put real effort, but can decide to make a gift, which reduces his/er own payoff. Nevertheless, this game is not zero-sum. For what it's worth, real-effort tasks are already established in experimental economics -- in works of Ariely e.g. \cite{arielysus}. Therefore, we refuse to draw any inferences from the behavior in the experiment to the behavior in real labor relations. The `employer' is renamed to originator and `employee' to follower. If the originator and the follower are both only interested in 
maximizing their payoff in a pure monetary case  and it is a one-shot game, the actual gift exchange will not take place.\\
\indent The experiment was conducted at University of Nottingham and consisted of one-shot games, whereby no subject participated twice. The participants were 20 years old in average and of both genders. Every one-shot game involves three players -- one originator and two followers. The originator has the choice to award none, one or both followers. The followers have four levels of rewarding (including non-rewarding) the originator. In the original game description, the originator and every follower have to give at least a minimum gift, which we denote non-gift for simplicity. At the beginning, the originator gets \pounds $8.3$ and every follower gets \pounds $11.1$. The additional payoff of the originator is the sum of margins from gift exchange with both followers. The additional payoff of every follower is the gift of the originator minus reduction through own gifts. The originator can give a fixed amount of \pounds$1.6$ to a follower. A follower can give \pounds $1$, \pounds $2$ or \pounds $3$, whereby 
his/er payoff reduces by \pounds $0.5$, \pounds $1$ or \pounds $1.5$ accordingly.\\
\indent We split the 3-players game into two 2-players games. Fig.\ref{gamestr} shows the 2-players game between an originator and a follower in extensive form. Extensive form is known in AI as game tree. The originator has to decide for two of such 2-players games. After the originator makes his choice, the followers make their choices either sequentially or simultaneously. Every follower can observe both of originators' decisions. In the sequential case, first follower's decision can be seen by the second follower. Besides mutual visibility, both 2-players games are independent. Adding both games, the originator's total payoff ranges between \pounds $5.1$ and \pounds $14.3$. The follower's total payoff ranges between \pounds $9.6$ and \pounds $12.7$.\\          
\section{Dataset}
\indent $123$ subjects participated in the game -- $84$ for the sequential case and $39$ for the simultaneous case. $\tfrac{123}{3}=41$ originators have made $41\times 2 = 82$ decisions -- two 2-players games per originator. The follower were asked to submit their decisions for every possible combination of others' observable decisions. There are $4$ decision combinations for an originator. First followers in the sequential case submitted $4*\tfrac{84}{3} = 112$ and all followers in the simultaneous case submitted $4*2*\tfrac{39}{3} = 104$ decisions. Second followers in the sequential case submitted $4*4*\tfrac{84}{3} = 448$ decisions. Therefore, we have a dataset of total $746$ human decisions.\\     
\section{Assumption of determinism}
\indent Modeling human behavior outside of game playing with human subjects should not be confused with prediction algorithms of artificial players. Quite the contrary, artificial players can manipulate the predictability of human subjects by own behavior. For instance, an artificial player, which always throws `stone' in roshambo, would success at predicting a human opponent always throwing `paper' in reaction. Otherwise, if an artificial player maximizes its payoff based on opponent modeling, it would face a change in human behavior and have to deal with it. This case is more complex than a spectator prediction model for an `only-humans' interaction. This work is restricted on modeling behavior without participating.\\
\indent Human behavior can be modeled as either deterministic or non-deterministic. Although human subjects fail at generating truly random sequences as demanded by mixed strategies equilibrium, non-deterministic models are especially used in case of artificial players in order to handle uncertainties.\\
\indent `Specifically, people are poor at being random and poor at learning optimal move probabilities because they are instead trying to detect and exploit sequential dependencies. ... After all, even if people don't process game information in the manner suggested by the game theory player model, it may still be the case that across time and across individuals, human game playing can legitimately be viewed as (pseudo) randomly emitting moves according to certain probabilities.' \cite{actrgames} In the addressed case of spectator prediction models, non-deterministic view can be regarded as too shallow, because deterministic models allow much more exact predictions. Non-deterministic models are only useful in cases, where a proper clarification of uncertainties is either impossible or costly. To remind, deterministic models should not be considered to obligatory have a formal logic shape.\\
\section{Nominal, Ordinal or Numeric}
\indent The usage of right data types is essential for correct data analysis. There are basically three categories, in which variables can be classified -- nominal, ordinal and numeric. Nominal variables assume values from a finite set, which has no order. Ordinal variables are like nominals plus ordering relationship over the set of values. Numeric variables assume real numbers $\mathbb{R}$ as values. Ordinal values can be projected into numeric under assumption about their distribution over the number axis. In contrast, nominal values can not.\\
\indent Some variables, which impact human actions, are actions of other players. Since presuming human preferences over the outcomes has no base, an ordering relationship over the actions can not be presumed as well. In the addressed problem, all variables impacting human actions are actions of others. Preferences over outcomes in the earlier described GEG can not be presumed. For instance, the outcomes $(.4,.6)$ and $(1.4,.1)$ (Fig.\ref{gamestr}) are the total payoffs $(\pounds 8.7,\pounds 11.7)$ and $(\pounds 9.7,\pounds 11.2)$. An egoistic follower would prefer the first and altruistic one the second. Since the variables have to be nominal and not even ordinal, they can not be projected into real numbers. An application of a linear model as in the original paper \cite{sefton} becomes therefore nonsense for this data.\\
\section{Results}
\begin{figure}
\includegraphics[scale=0.5]{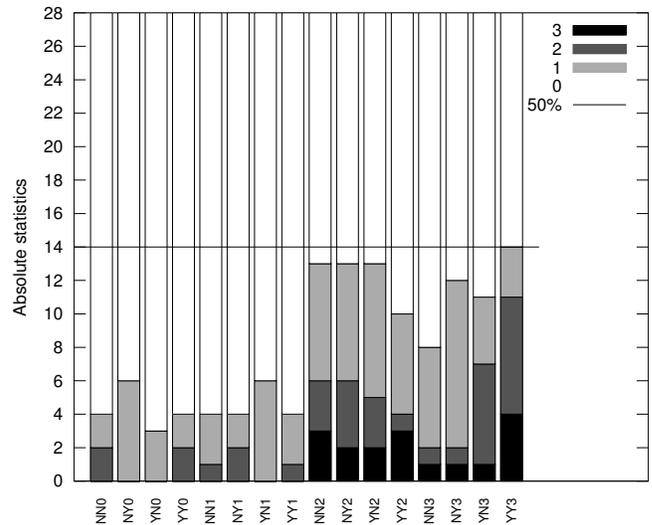}
\caption{Choice of the second follower -- absolute statistics depending on other players' decisions encoded on x-axis as: other follower's received, own received and other follower's given. N is \pounds0 and Y \pounds1.6.}
\label{secondf}
\end{figure}

\indent Originator's both decisions are nominal or rather boolean -- it is either a gift or not. In average, originators gift in $36.6$\% of samples. We calculate Kappa \cite{cohen60,fleiss} to measure the inter-rater agreement between these two decisions. Having zero Kappa as null hypothesis, the significance of the measured Kappa can be calculated. Tab.\ref{origtab} displays significant fair agreement between originator's both decisions in the sequential case. There is no significant agreement between them in the simultaneous case. Unfortunately, the data is too marginal and Fisher's test \cite{fisher} does not show any significant difference between the frequencies in both cases -- p-value is $0.4686$. We can at least claim that both decisions are dependent in the sequential case.\\
\begin{table}
\begin{center}
\caption{Originator's two decisions -- absolute statistics and agreements.}\label{origtab}
\begin{tabular}{|c|c|c|c|}
\hline
 & Sequential & Simultaneous & Sum\\
\hline
Gifts for 1st  &  $9$  & $6$  &  $15$  \\
Gifts for 2nd  &  $10$  & $5$  &  $15$  \\
All samples     & $28$ & $13$  & $41$\\
\hline
Kappa  & $.4432$ & $.2169$ & $.3692$\\
p-value  & $.017$ & $.22$ & $.014$\\ 
\hline 
\end{tabular} 
\end{center} 
\end{table}
\indent Tab.\ref{follabs} shows absolute statistics for the follower's decisions, which did not observe another follower. Besides own received gift, there is the gift received by the other follower, which might have an impact on the observing follower's decision. If no own gift is received in the simultaneous case, Fisher's test results a p-value of $0.0496$ for gifting $>$\pounds$0$ depending on whether or not the other follower received a gift. Receiving less than the other follower is therefore significantly reciprocated in the simultaneous case only. The significance of this result is thoroughly discussed in section \ref{phacking}. In the sequential case, there is no significant difference between decision frequencies depending on the other's received gift. Obviously, the order between the follower delivers a reason for an unequal treatment.\\ 
\indent Tab.\ref{follkap} lists agreements as well as their significances between subsets of own decisions and observed decisions. The subsets of own decisions are defined by thresholds on gift size. We use thresholds to define subsets, because based on decreasing frequencies by raising gift size (Tab.\ref{follabs}), an order on the gift decisions can be derived. One can see that only own received gift has a significant influence on own decision in sequential as well as in simultaneous cases. Since only one variable has influence on the decision, the deterministic model is trivial -- gift $>$\pounds$0$ in the sequential case having received a gift, non-gift anywhere else.\\
\begin{table}
\begin{center}
\caption{Follower's decision without observing another follower's decision -- absolute statistics.}\label{follabs}
\begin{tabular}{|c|c|cccc|cccc|}
\hline
\multicolumn{2}{|r|}{} & \multicolumn{4}{c|}{Sequential} & \multicolumn{4}{c|}{Simultaneous} \\
\hline
\multicolumn{2}{|r|}{Own gift} & \pounds$0$ & \pounds$1$ & \pounds$2$ & \pounds$3$ &  \pounds$0$ & \pounds$1$ & \pounds$2$ & \pounds$3$ \\
\hline
\multicolumn{2}{|l|}{Received gifts} &       & & &               &       & & &   \\
Own & Other's &       & & &               &       & & &   \\
\hline
\pounds $0$ & \pounds$0$ &  $21$ &   $6$ &  $1$ & $0$ & $19$ &   $7$  &  $0$  &  $0$ \\
\pounds $0$ & \pounds $1.6$ &   $21$ &   $6$ &   $1$ &   $0$ & $25$ &   $0$ &   $1$ &   $0$ \\
\pounds $1.6$ & \pounds $0$ &    $13$ &   $6$ &   $5$ &   $4$  & $16$  &  $6$  &  $4$  &  $0$ \\
\pounds $1.6$ & \pounds $1.6$ &   $13$ &  $6$ &   $4$ &   $5$ & $16$ &  $3$ &   $4$  &  $3$ \\
\hline
\end{tabular} 
\end{center} 
\end{table}
\begin{table}
\begin{center}
\caption{Follower's decision without observing another follower's decision -- agreements with originator's decisions.}\label{follkap}
\begin{tabular}{|c|cc|cc|}
\hline
 & \multicolumn{2}{c}{\textcolor{white}{.......}Sequential\textcolor{white}{.......}} & \multicolumn{2}{c|}{Simultaneous}\\
 & Kappa  & p-value  & Kappa    &  p-value           \\
\hline
Own gift ... & \multicolumn{4}{c|}{ ... vs. received gift} \\
\hline
$>$\pounds $0$   & $.2857$ & $.0012$   & $.2308$ & $.0093$  \\
$>$\pounds $1$   & $.2857$ & $.0012$   & $.1923$ & $.0249$  \\
$>$\pounds $2$   & $.1607$ & $.045$    & $.0577$ & $.2781$  \\
\hline
& \multicolumn{4}{c|}{... vs. other's received gift} \\
\hline
$>$\pounds $0$   & $0$ & $.5$          &   & \\
$<$\pounds 1   &       &         & $.1154$ & $.1197$  \\
$>$\pounds 1   & $0$ & $.5$          & $.0769$ & $.2164$ \\
$>$\pounds 2   & $.0179$ & $.4251$   & $.0577$ & $.2781$  \\
\hline 
\end{tabular} 
\end{center} 
\end{table}
\indent As Fig.\ref{secondf} shows, non-gift covers $\geq50$\% of decisions for the second follower for all possible combinations of input variables and 71.2\% in average. One can of cause assume that some hidden variables influence the gift decision. Since we do not see these variables, we can not build a useful valid deterministic model -- none can be better than the null hypothesis suggesting non-gift everywhere. We restrict the analysis to agreements between the decision and the three observed variables.\\
\begin{table}
\begin{center}
\caption{Second follower's decision -- agreements.}\label{sfollkap}
\begin{tabular}{|c|cc|}
\hline
& \multicolumn{2}{c|}{Sequential case} \\
& Kappa &  p-value \\
\hline
Own gift ... & \multicolumn{2}{c|}{ ... vs. received gift} \\
\hline
$>$\pounds $0$   & $.0223$ & $.3183$\\
$>$\pounds $1$   & $.0223$ & $.3183$\\
$>$\pounds $2$   &  $.0134$ & $.3884$\\
\hline
& \multicolumn{2}{c|}{... vs. other's received gift} \\
\hline
$>$\pounds $0$   & $.0045$ & $.4624$ \\
$>$\pounds $1$   & $.0402$ & $.1975$ \\
$>$\pounds $2$   & $.0134$ & $.3884$ \\
\hline 
& \multicolumn{2}{c|}{... vs. other's gift} \\
\hline
   &  $.0446$   &  $.0509$\\
\hline 
\end{tabular} 
\end{center} 
\end{table}
\indent Tab.\ref{sfollkap} shows the agreements between subsets of the second follower's decisions and the decisions of the originator -- none of them is significant. It also shows the agreement between the decisions of the first and the second follower, whereby correspondence between the values' sets of both variables is assumed. This agreement is not significant. Therefore, the value sets of both variables have to be transformed. Tab.\ref{fsdeepkap} shows agreements between both variables transformed to booleans in different combinations. The highest agreement and the lowest p-value are achieved for the first follower's gift $>$\pounds$1$ and the second one's $>$\pounds$0$. Once the first follower is extra generous, the second one is also driven to gift the originator.\\    
\begin{table}
\begin{center}
\caption{First and second follower's decision -- agreements between subsets.}\label{fsdeepkap}
\begin{tabular}{|c|cc|cc|cc|}
\hline
1st follower & \multicolumn{2}{c}{$>$\pounds0} & \multicolumn{2}{c}{$>$\pounds1} & \multicolumn{2}{c|}{$>$\pounds2}\\
 & Kappa &  p-value & Kappa &  p-value & Kappa &  p-value\\
2nd follower &  &  &  &  & & \\
\hline
$>$\pounds$0$ &  $.1123$ &  $.0016$  & $.2634$ & $1.238e-8$ & $.1445$ & $.0068$ \\
\hline
$>$\pounds$1$ & $.0564$ & $.0365$    & $.1563$ & $.0005$ & $.1345$ & $.029$ \\
\hline
$>$\pounds$2$ & $.026$ & $.1826$      & $.0759$ &  $.0541$ & $.0456$ & $.2789$ \\
\hline
\end{tabular} 
\end{center} 
\end{table}
\indent To summarize, the frequency of non-gift is $63.4$\% for the originator, $60.7$\% for the first sequential follower, $73$\% for the simultaneous follower and $71.2$\% for the second sequential follower. According to Fisher test, none of these frequencies significantly deviates from the rest. The average non-gift frequency is $69$\%.\\  
\section{On p-hacking}
\label{phacking}
\indent It is not a secret that p-values closely under $0.05$ cause suspicion about the scientific methods used in research \cite{nuzzo}. Although p-value was never thought to be an objective criterion for proof or disproof of a hypothesis, many researchers misunderstand it and conduct the so called 'p-hacking' on the data to archive significant results.\\
\indent The results achieved through p-hacking might not be reproducible, since the a-priori probabilities of the hypotheses have to be incorporated as well. For instance, if the hypothesis is a long shot and has an a-priori probability of $5$\%, a p-value of $0.01$ raises the chance of its validity to only $30$\%. The more hypotheses are tested on p-value, the higher the probability to achieve a p-value under $0.05$. Obviously, the difference between long shots and good bets has to be derived from the researcher's expert knowledge, which is known to be absent in the case of an pure data scientist analyzing human data. Here, we suggest the data scientist to be 'agnostic' and use some background knowledge to advocate the result.\\
\indent During the data analysis in this paper, we got a p-value of $0.0496$ for the hypothesis that an unfairly treated simultaneous follower negatively reciprocates. Using the background knowledge about human reaction to unfairness \cite{chickensym}, we can assume that it is a good bet. If the a-priori probability for a good bet is assumed to be about $90$\%, a p-value of $0.05$\% raises its chance of validity to $96$\%.\\  
\section{Future work}
\indent During the work on this paper, we confronted the time consuming requesting, selection and reformatting of data. Unfortunately, there is no online portal, where most of the datasets are offered in a common format. This is an issue, which we will address in the future. Like in the field of bioinformatics, common formats are an important part of an interdisciplinary research infrastructure and are needed to accelerate the progress \cite{forsinfr}.\\
\indent As for methodological aspects of Machine Learning in the context of Experimental Economics, we would like to use the advanced pattern mining techniques for economic game data analyses. For example, in papers \cite{Liris-6329,Baixeries:2013} was made an attempt to use sequential patterns and similarity dependencies on pattern structures for video game players' behavior analysis, in particular sequential attribute dependencies might be a tool of choice. We will try to apply sequential pattern mining in a supervised task, where the outcome of a game (or a turn) is a target attribute \cite{Kuznetsov:2004,Buzmakov:2013} to see which patterns better generalize the user behavior. These experiments are able not only to broad the tools of experimental economics, but also help to reveal potentially new knowledge of human behavior in games based on sequential pattern description.\\    
\section{Conclusion}
\indent First of all, the average non-gift frequency is only $69$\% in the studied one shot GEG. These are far away from the $100$\%, which an egoistic payoff maximization assumption would predict. But, it is also over $50$\% in almost all cases. There, it is impossible to create valid nontrivial deterministic models of human behavior without having access to the hidden variables, which determine the choice. Only if the first follower receives a gift in the sequential case, the frequency of gifts goes slightly over $50$\%.\\
\indent Although first follower's decision depends only on his received gift and second follower's decision does not depend on originators' decisions at all, originators' decisions are interdependent in the sequential case. The order between the players obviously delivers a reason for the first follower to not mind differences in gifts. Having no order in the simultaneous case leads to significant negative reciprocation of receiving less than the second follower.\\   
\indent A curious finding is that not minimal but extra generosity is 'contagious' for the followers. Second follower reacts only on the first follower being extra generous and with normal generosity.\\
\section*{Acknowledgment}
\indent The authors would like to thank Martin Sefton for the friendly reply and the transfer of data. We also thank the people, who provided the Weka library \cite{weka}, for their wonderful work, as well as Minato Nakazawa for the fmsb package.\\
\newpage

\bibliographystyle{IEEEtran}
\bibliography{human05}
\end{document}